\documentclass[10pt,twocolumn,letterpaper]{article}

 \usepackage{cvpr}              
\definecolor{cvprblue}{rgb}{0.21,0.49,0.74}
\usepackage[pagebackref,breaklinks,colorlinks,allcolors=cvprblue]{hyperref}

\definecolor{forestgreen}{rgb}{0.0, 0.5, 0.0}
\definecolor{ashgrey}{rgb}{0.7, 0.75, 0.71}

\usepackage{tikz}
\usepackage{tcolorbox}
\usepackage{pifont}
\usepackage{bbding}
\usepackage{graphicx}    
\usepackage{amsmath}     
\usepackage{hyperref}    
\usepackage{multirow}    
\usepackage{xcolor}      
\usepackage[table]{xcolor} 
\usepackage{adjustbox}
\definecolor{color_gray}{RGB}{229,229,229}
\definecolor{color_blue}{RGB}{252,182,165}
\definecolor{color_pink}{RGB}{255,217,178}
\definecolor{color_yellow}{RGB}{255,255,204}
\definecolor{secondcolor}{RGB}{189,215,238}
\definecolor{firstcolor}{RGB}{255,153,153}
\newcommand{\firstcolor}[1]{\cellcolor[rgb]{1,.60,.60}{#1}}
\newcommand{\secondcolor}[1]{\cellcolor[rgb]{.741,.843,.933}{#1}}
\definecolor{bb}{rgb}{0.0, 0.0, 0.5}
\definecolor{Gray}{gray}{0.9}
\usepackage{adjustbox}
\usepackage{multirow}
\usepackage{colortbl}
\usepackage{tcolorbox}
\newtcolorbox{commentbox}{
    colback=gray!2,          
    colframe=black,         
    arc=2pt,                  
    boxrule=0.5pt,            
    left=2pt,                 
    right=2pt,                
    top=2pt,                  
    bottom=2pt,               
    boxsep=1pt,               
    fontupper=\itshape,       
}

\def\paperID{9185} 
\def\confName{CVPR}
\def\confYear{2026}

\title{Towards an Incremental Unified Multimodal Anomaly Detection: Augmenting Multimodal Denoising From  an Information Bottleneck Perspective}

\author{Kaifang Long\textsuperscript{\rm 1}, \quad Lianbo Ma\textsuperscript{\rm 1\thanks{Corresponding author.}}, \quad Jiaqi Liu\textsuperscript{\rm 2}, \quad Liming Liu\textsuperscript{\rm 1}, \quad Guoyang Xie\textsuperscript{\rm 3}\\
\textsuperscript{\rm 1}Software College, Northeastern University, China, \textsuperscript{\rm 2}Independent Researcher, \textsuperscript{\rm 3}CATL, China\\
{\tt\small longkf@stumail.neu.edu.cn, malb@swc.neu.edu.cn, guoyang.xie@ieee.org}
}

\begin{document}
\maketitle
\begin{abstract}
The quest for incremental unified multimodal anomaly detection seeks to empower a single model with the ability to systematically detect anomalies across all categories and support incremental learning to accommodate emerging objects/categories. Central to this pursuit is resolving the catastrophic forgetting dilemma, which involves acquiring new knowledge while preserving prior learned knowledge. Despite some efforts to address this dilemma, a key oversight persists: ignoring the potential impact of spurious and redundant features on catastrophic forgetting. In this paper, we delve into the negative effect of spurious and redundant features on this dilemma in incremental unified frameworks, and reveal that under similar conditions, the multimodal framework developed by naive aggregation of unimodal architectures is more prone to forgetting. To address this issue, we introduce a novel denoising framework called IB-IUMAD, which exploits the complementary benefits of the Mamba decoder and information bottleneck fusion module:  the former dedicated to disentangle inter-object feature coupling, preventing spurious feature interference between objects; the latter serves to filter out redundant features from the fused features, thus explicitly preserving discriminative information. A series of theoretical analyses and experiments on MVTec 3D-AD and Eyecandies datasets demonstrates the effectiveness and competitive performance of IB-IUMAD. \href{https://github.com/longkaifang/IB-IUMAD}{Code Link}.
\end{abstract}
\section{Introduction}
\label{sec:intro}
\textbf{Motivation.} Multimodal anomaly detection (MAD) has emerged as a pivotal task in industrial quality inspection, with the aim of recognizing and locating product surface defects using RGB and depth images \cite{DBLP:conf/aaai/GuZLCPGJSWM24,li2025multi,long2025}.  Most existing MAD methods follow an \textit{N-objects-N-models} principle \cite{costanzino2024multimodal}, where a separate model is designed for each object/category (i.e., an industrial product). However, this paradigm suffers from significant limitations of high computational cost, substantial memory consumption, and weak generalization. Then, the \textit{N-objects-One-model} has become a promising schema, enabling a single model to detect anomalies between multiple categories \cite{DBLP:conf/ijcai/ChengGZ0DW25,guo2025dinomaly}. Especially, in real industrial scenarios, the continuous emergence of new objects is a prevalent situation \cite{DBLP:conf/aaai/0004WNCGLWWZ24}, which requires MAD models based on the \textit{N-objects-One-model} to support incremental learning for the adaptation to new ones. Thus, this study focuses on the exploration of the incremental unified multimodal anomaly detection (IUMAD) task.

\begin{figure}[t!]
\centering
\includegraphics[scale=0.36]{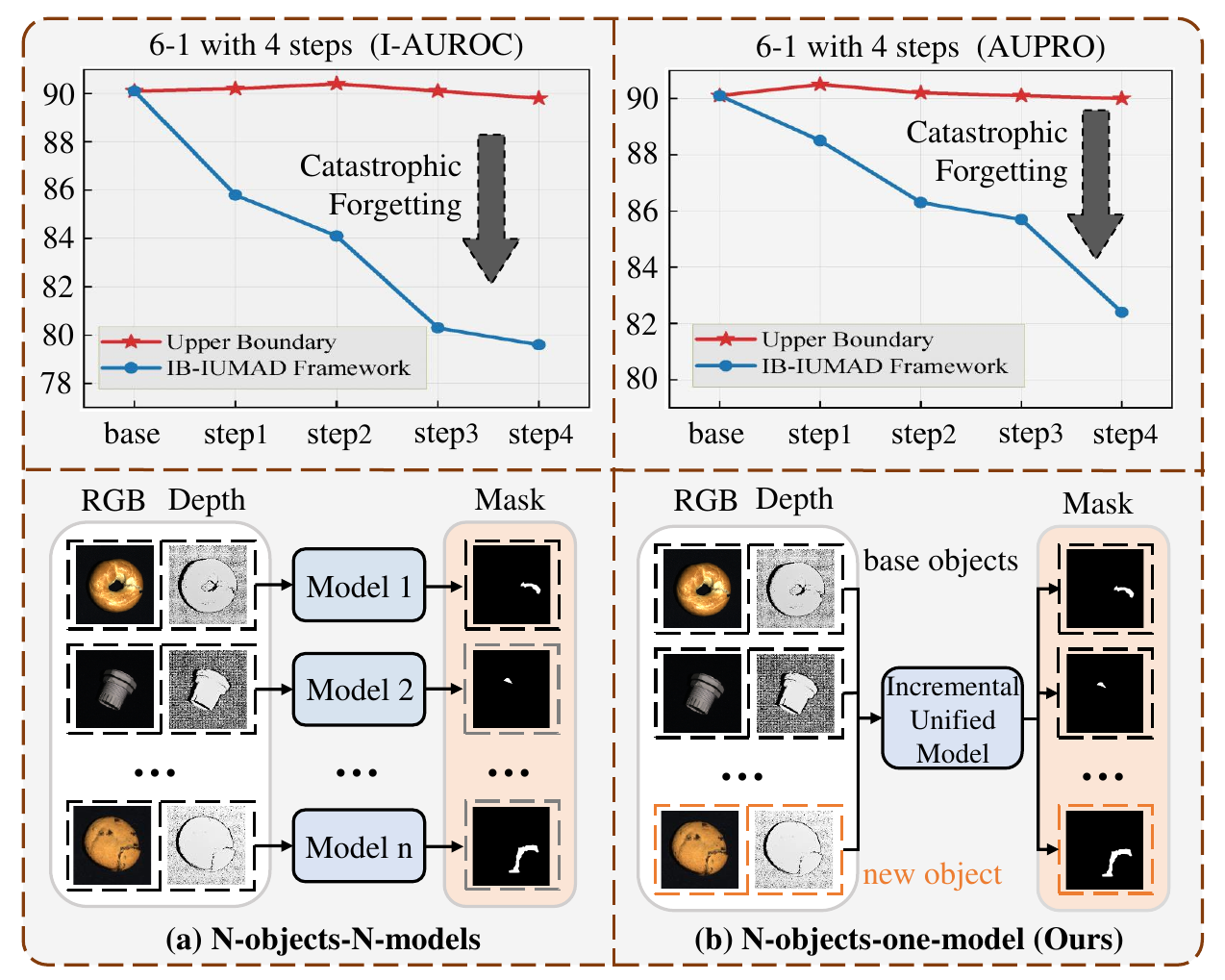}
  \caption{\textbf{Top} shows the performance of IUMAD task on the MVTec 3D-AD dataset, where catastrophic forgetting occurs significantly. \textbf{\textit{Bottom}} is the comparison of our incremental unified multimodal framework with previous paradigms.}
 \label{figure1_121}
\end{figure}


\noindent\textbf{Challenges.} In fact, existing works about \textit{N-objects-One-} \textit{model} focus on unimodal anomaly detection (e.g., for RGB images) \cite{li2022towards} rather than IUMAD. As shown in Figure \ref{figure1_121}, the performance on the base unified model, trained with 6 objects in MVTec 3D-AD, is excellent but decreases sharply as the model trains on new objects incrementally. Thus, central to this quest is how to effectively preserve acquired knowledge as new objects are continuously introduced, i.e., how to overcome catastrophic forgetting dilemma. For this purpose, the object-aware self-attention and semantic compression loss are used to capture semantic boundaries, improving adaptability to new objects \cite{tang2024incremental}. Then, the gradient projection method is devised for stable incremental learning using diffusion models \cite{li2025one}. However, these works neglect the potential impacts of inter-object spurious features and redundant information in fused features towards catastrophic forgetting, which has been demonstrated in conventional machine learning scenarios \cite{fang2024dynamic}. More critically, we observe that compared to unimodal detection scenarios, the aforementioned impacts exert substantially amplified influence on MAD due to the inherent complexity of cross-modal feature fusion. Consequently, the incremental unified design for MAD presents significantly greater technical challenges, particularly in maintaining inter-modal interactions during sequential learning phases.

To investigate the above issue, we conducted a series of preliminary experiments on IUF \cite{tang2024incremental} and its variant to examine the impact of spurious and redundant features on catastrophic forgetting in incremental unified unimodal and multimodal frameworks, evaluating accuracy and average forgetting when new objects are learned incrementally. As shown in Figure \ref{Fig_2_sec2}, introducing spurious and redundant feature perturbations leads to substantial performance degradation in both frameworks, along with a marked increase in average forgetting. Especially for multimodal frameworks, the feature fusion process increases the model’s susceptibility to capturing spurious and redundant features from different modalities, resulting in more severe deterioration than unimodal and even causing performance collapse.

\textbf{How to mitigate catastrophic forgetting triggered by spurious and redundant features in IUMAD?} A straightforward and intuitive solution is to devise two independent modules with distinct functions: one dedicated to mitigating interference from inter-object spurious features, and the other focused on filtering out redundant information in multimodal fusion features. Building on these insights, we extend the Mamba technique for IUMAD and enhance multimodal fusion from the information bottleneck perspective, thereby proposing a novel denoising framework, called IB-IUMAD, which leverages the complementary strengths of the Mamba decoder and information bottleneck fusion module. Specifically,  the former focuses on integrating the fine-grained features with label information into the multimodal reconstruction network, effectively disentangling inter-object feature coupling and preventing spurious feature interference. The latter is designed to filter out redundant information from the fused multimodal features, explicitly preserving discriminative features. Extensive experiments show that IB-IUMAD consistently improved performance compared to baselines.

\textbf{Contributions.} In summary, the main contributions of this work are as follows:
\begin{itemize}
    \item We empirically validate the impact of spurious and redundant features on catastrophic forgetting and then identify the critical denoising module designs that boost IUMAD performance from an information bottleneck perspective.

    \item We propose a novel framework, called IB-IUMAD, which leverages the complementary strengths of the Mamba decoder and information bottleneck regularization to mitigate the interference of spurious and redundant features. 

  \item  A series of theoretical analyses and experiments demonstrates that the proposed method consistently improves performance in terms of accuracy, memory usage, and frame rate.
\end{itemize}

To the best of our knowledge, this is the first work to address MAD in an incremental and unified manner. Our method outperforms state-of-the-art baselines across four incremental settings, achieving higher accuracy and frame rates with lower forgetting and memory costs. Specifically, under the 6-1 with 4-step setting, IB-IUMAD boosts I-AUROC/AUPRO by 3.5\%/2.9\% and reduces forgetting by 5.8\%/1.5\% on MVTec 3D-AD. In the 10-0 with 0-step setting, it cuts memory usage by 44× and delivers at least 41× faster inference than the N-objects-N-models approach, while maintaining comparable performance.
\section{Spurious and Redundant Features Impact}
\label{Scrutiniz}
This section investigates the impact of spurious and redundant features on catastrophic forgetting in incremental unified frameworks, with a focus on multimodal settings.

\subsection{Problem Formulation}
Figure \ref{figure1_121}-(b) outlines the formal definition of the problem setting for IUMAD tasks, which aim to develop a single model capable of: comprehensive anomaly identification across heterogeneous objects, and continuous knowledge integration through incremental learning to accommodate emerging object types. Specifically, we divide the dataset containing $N$ objects into $T$ training steps ($T \leq N$). In the $t$-th training step, the model updates its weight parameters solely based on the current object, without accessing any data from previous steps. During testing at the $t$-th step, we evaluate anomaly scores for all seen objects so far and employ the forgetting metric (FM) to quantify the degree of forgetting of previously learned knowledge. For example, in the 6-1 with 4 steps or 6-4 with 1 step setting, we first train a base model with six objects, and then incrementally learn one or four additional objects per step across four or one incremental step(s), respectively, to obtain the final IUMAD model. Other settings follow the same principle.

\subsection{Empirical Study Settings}
\textbf{Benchmark.}\quad  IUF \cite{tang2024incremental} is selected as the foundational framework, given its widespread application in existing research \cite{lee2025continual}. Our primary goal is to examine how spurious and redundant features exacerbate catastrophic forgetting in incremental settings, evaluating three input patterns: RGB-only, depth-only, and their combination.

\textbf{Evaluation Metrics.}\quad We evaluate model performance using several standard metrics (I-AUROC, P-AUROC, and AUPRO) and introduce the forgetting metric (FM) to quantify the degree of forgetting of previously learned knowledge when learning new objects incrementally; the lower the FM, the better the performance. FM as defined by:
\begin{equation}
FM = \frac{1}{N-1} \sum_{o}^{N-1} max_{s\in[1,N-1]} (I_{s,o}^{Acc}-I_{N,o}^{Acc}),
\end{equation}
where $N$ denotes the total incremental steps. $I_{s,o}^{Acc}/ I_{N,o}^{Acc}$ represents the I-AUROC, P-AUROC, or AUPRO score for the $o$-th object at the $s$-th$/N$-th step. $max_s(\cdot)$ denotes the maximum score drop for each object across all steps.

\textbf{Implementation Details.}\quad In our empirical study protocol, we train a unified model on 6 basic objects from  MVTec 3D-AD, followed by incrementally introducing the remaining objects over 4 steps (i.e., 6-1 with 4 steps), and simulate the interference of spurious and redundant features by injecting background information (acquired from the other objects, as detailed in Supplementary Material) and Berlin noise with different intensities into the model. To ensure statistical robustness, we perform four runs with different random seeds and report the mean and standard deviation of results. Each run consists of: (i) initial training for 1,000 epochs on basic objects, and (ii) training for 800 epochs per incremental step. 

\subsection{Empirical Observations}
\begin{figure}[t!]
\centering
\begin{minipage}{0.98\linewidth}
		\centerline{\includegraphics[width=\textwidth]{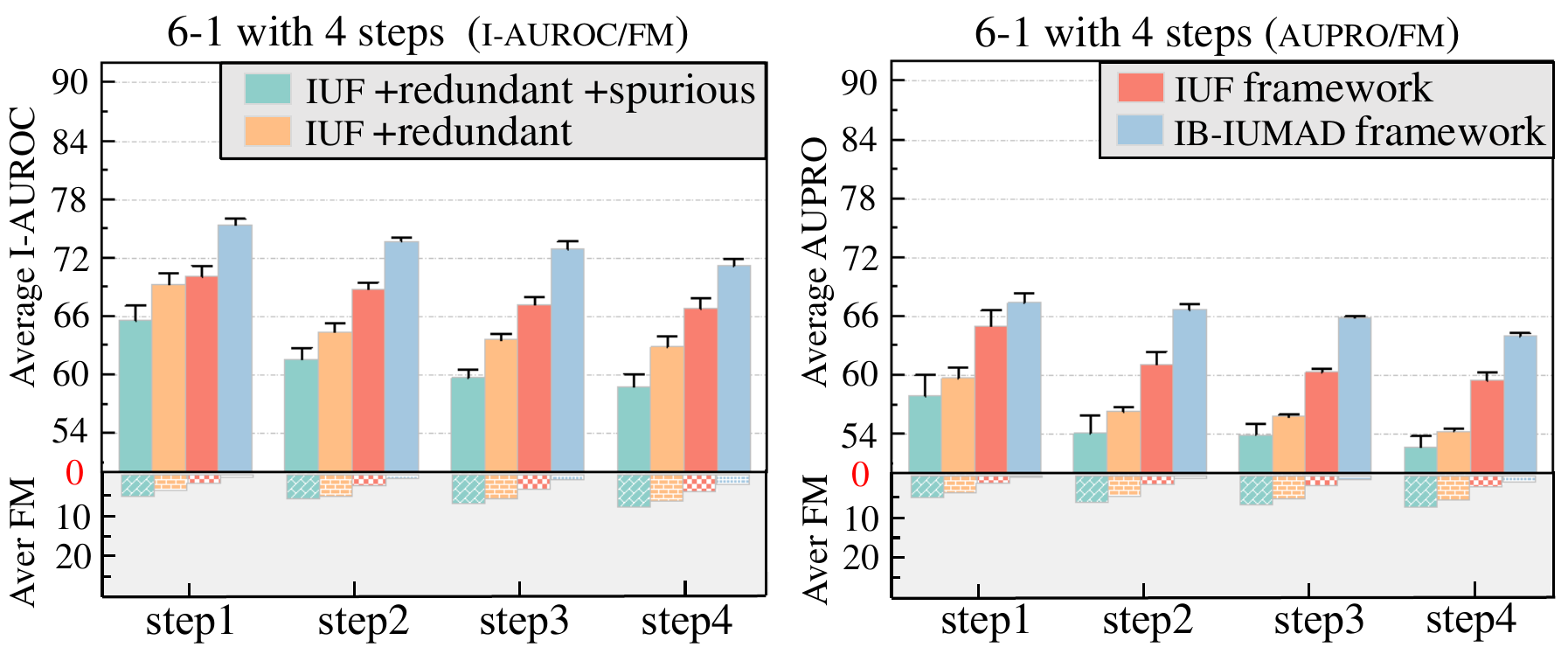}}
	\end{minipage}
 \centerline{(a) Comparison results under the Depth-only setting.}
\begin{minipage}{0.98\linewidth}
		\centerline{\includegraphics[width=\textwidth]{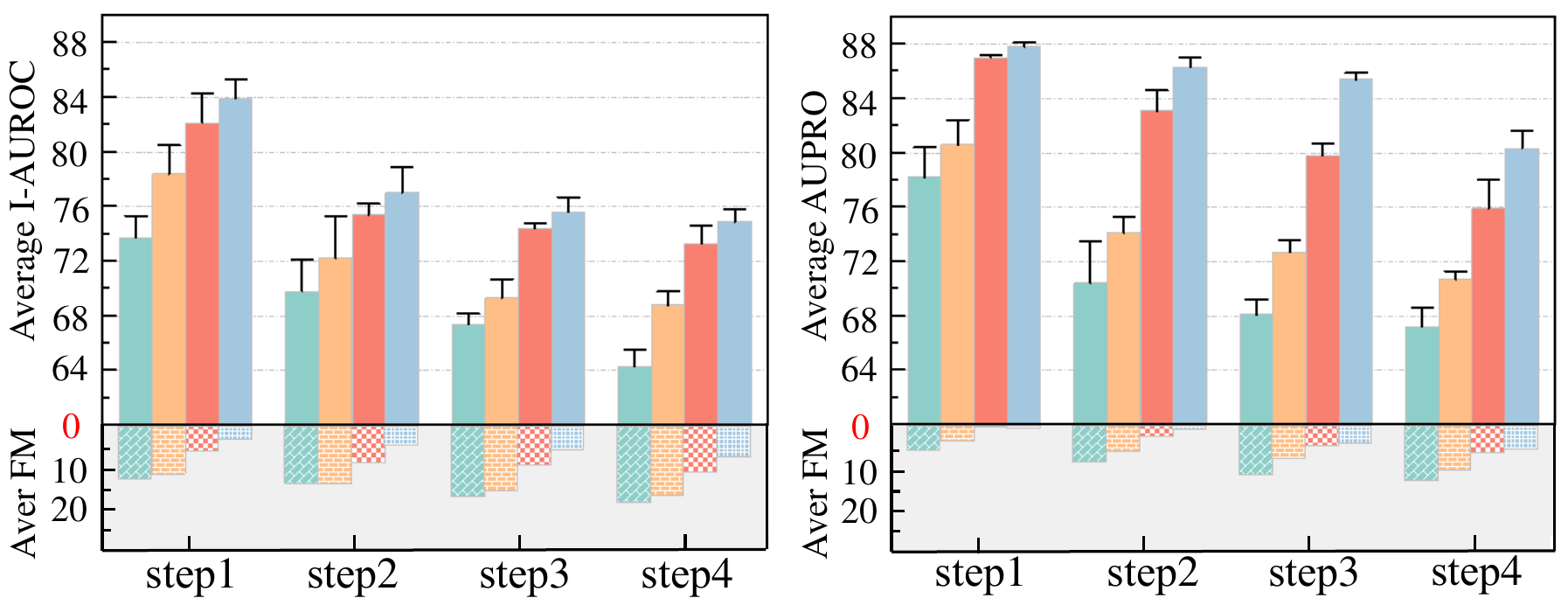}}
	\end{minipage}
  \centerline{(b) Comparison results under the RGB-only setting.}
\begin{minipage}{0.98\linewidth}
		\centerline{\includegraphics[width=\textwidth]{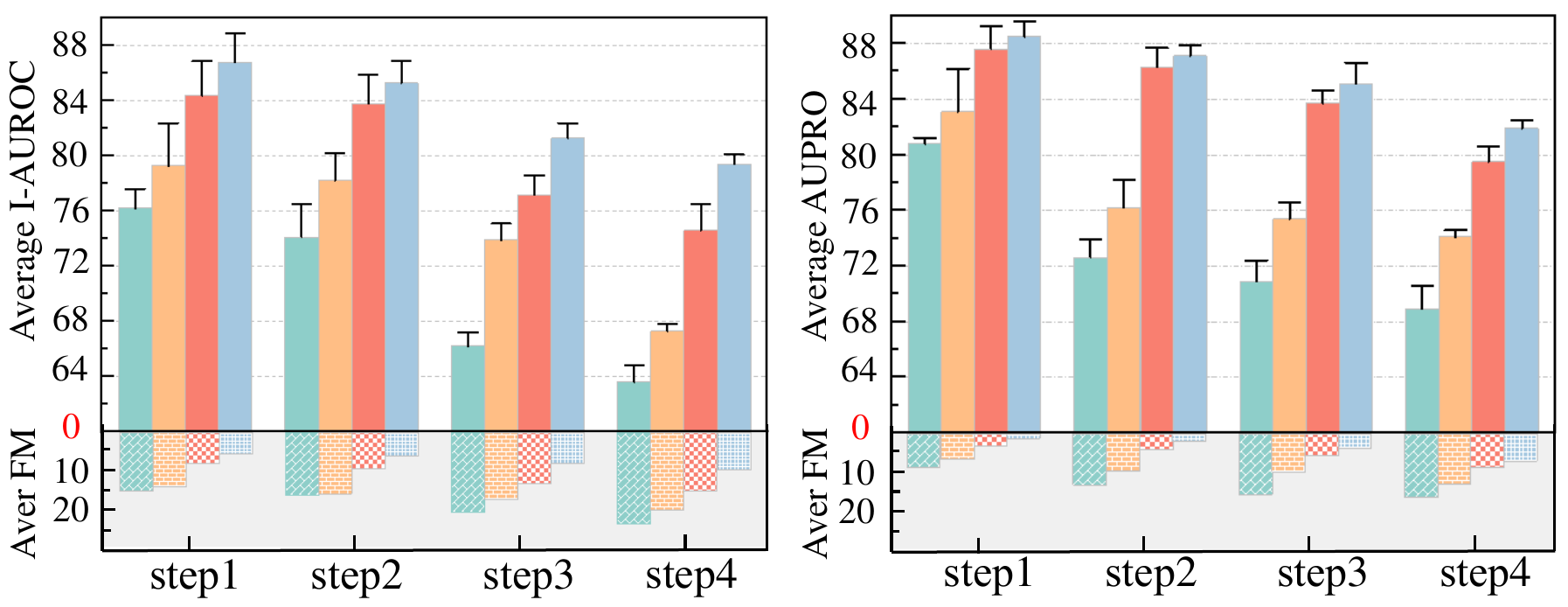}}
	\end{minipage}
  \centerline{(c) Comparison results under the multimodal setting.}
	\caption{The impact of spurious and redundant features on catastrophic forgetting in incremental unified frameworks.}
  \label{Fig_2_sec2}
\end{figure}

To evaluate the impact of spurious and redundant features on catastrophic forgetting in the incremental unified unimodal and multimodal framework, under the 6-1 with 4 steps setting, we report the performance at each incremental step, involving I-AUROC, AUPRO, and FM. In brief, as previewed in Figure \ref{Fig_2_sec2}, we derive the following \underline{\textbf{observations}}: 

\ding{182} Under the unimodal setting, injecting redundant features degrades model performance and accelerates forgetting, while introducing additional spurious features further intensifies this decline. This indicates that the model becomes increasingly sensitive to such features during incremental learning, thereby amplifying their negative impact on overall performance. (Figure \ref{Fig_2_sec2}-a and b)

\ding{183} Under the multimodal setting, we observe a performance degradation trend similar to the unimodal case, but with a more pronounced decline. A key contributing factor is the inherent complexity of cross-modal fusion, increases the likelihood of capturing spurious or redundant features, thereby exacerbating catastrophic forgetting. (Figure \ref{Fig_2_sec2}-c)

\ding{184} As shown in Figure \ref{Fig_2_sec2} (blue bars), employing a denoising constraint for IUMAD, such as information bottleneck regularization, can effectively alleviate the negative impact of spurious and redundant features on overall performance.

\ding{185}  In summary, how to effectively mitigate the catastrophic forgetting problem triggered by spurious and redundant features is crucial to improving model performance.

Based on the above insights, we identify that spurious and redundant features pose a considerable challenge to IUMAD tasks. This motivates us to design an incremental unified denoising framework for improving the model's performance.
\section{Method}
In this section, we present IB-IUMAD, a denoising method that mitigates catastrophic forgetting arising from spurious and redundant features. We further outline its framework, detailing the denoising module and theoretical analysis.

\begin{figure*}[t]
  \centering
  \setlength{\abovecaptionskip}{0.1cm}
      \includegraphics[scale=0.53]{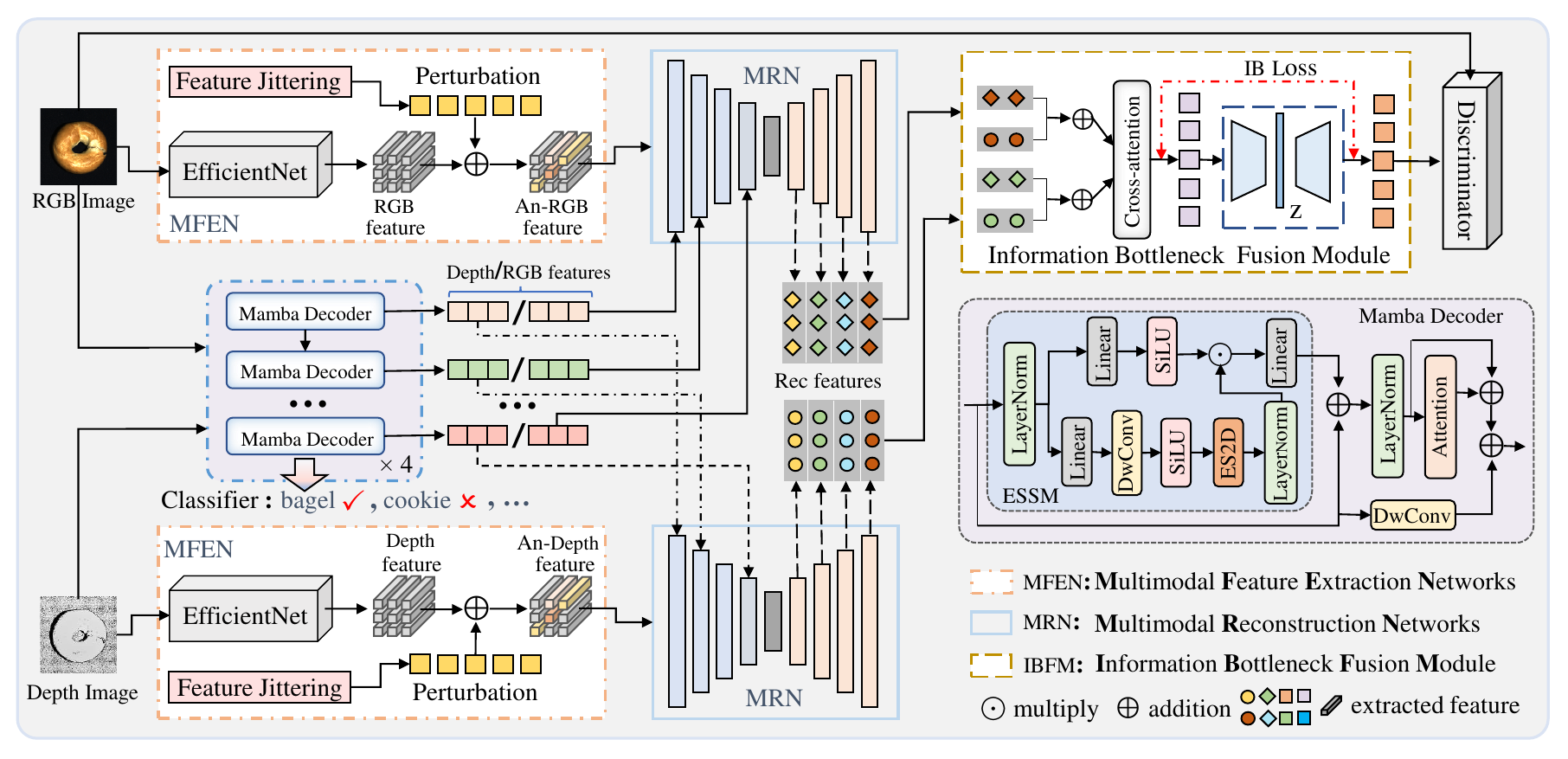}
  \caption{The overall framework of IB-IUMAD, where the Mamba decoder and information bottleneck fusion module (IBFM) are the core designs of this work, aims to mitigate the effects of spurious and redundant features on catastrophic forgetting.}
  \label{entire:3}
\end{figure*}

\subsection{Overview}
Figure \ref{entire:3} shows the skeleton of IB-IUMAD, which comprises Multimodal Feature Extraction Networks (MFEN), Mamba decoders, Multimodal Reconstruction Networks (MRN), an Information Bottleneck Fusion Module (IBFM), and a Discriminator. MFEN aims to extract RGB and depth image features via EfficientNet and synthesize them into abnormal features using feature jitters. Then, MRN restores these abnormal features to normal ones while acquiring multi-scale multimodal representations. To mitigate interference of spurious features between objects during normal feature reconstruction within the MRN, we employ the Mamba decoder to disentangle inter-object feature coupling.  Subsequently, IBFM serves to fuse multi-scale multimodal features from MRN and imposes information bottleneck regularization to filter out redundant information from the multimodal fusion features,  which effectively alleviates catastrophic forgetting. Particularly, the Mamba decoders and IBFM constitute the core components of the IB-IUMAD framework. The following subsections detail their designs.

\subsection{Mamba Decoder}
The above experiments demonstrate that spurious and redundant features significantly exacerbate catastrophic forgetting in IUMAD tasks. To improve its performance, it is crucial to enhance the model's denoising capability to mitigate the interference of spurious and redundant features.

Specifically, the generation of spurious features primarily stems from the tight coupling of features between objects that share the same feature space.  Especially,  as the model incrementally learns new objects, it tends to indiscriminately update the feature space of previously learned ones, ultimately making the MRN more vulnerable to inter-object spurious features during reconstruction. To address this issue, inspired by \cite{pei2025efficientvmamba}, we incorporate the Mamba decoders and a label classifier to disentangle inter-object feature coupling. As shown in Figure \ref{entire:3}, each Mamba decoder comprises an efficient state space module (ESSM), deep separable convolution (DwConv), and an attention mechanism. Firstly, the ESSM employs DwConv and Efficient 2D Scanning (ES2D) \cite{long2025enhancing} to efficiently sample each visual patch, extracting fine-grained information from RGB and depth images. Then, the extracted features are processed by the attention mechanism, forming the Mamba decoder's output features, as follows:
\begin{equation}
\begin{split}
\hat{X}_{R/D}^{i+1}&=DwConv(X_{R/D}^i),\\
\widetilde{X}_{R/D}^{i+1}&=ESSM(LN(X_{R/D}^i)),\\
X_{R/D}^{i+1}&=Attention(LN(\widetilde{X}_{R/D}^{i+1}))+\hat{X}_{R/D}^{i+1},\\
\end{split}
  \label{equ:21_111}
\end{equation}
where $R/D$ denotes RGB and depth images, respectively; $X_{R/D}^i$ represents the output of the $i$-th mamba decoder, $i\in[0,4]$; specifically, when $i=0$, $X_{R/D}^i$ denotes the original image features. $LN$ indicates layer normalization.

Subsequently, the output of each Mamba decoder is independently integrated into the MRN to assist feature reconstruction, while the output of the final block is also fed into the classifier for label prediction.  This design aims to guide the model to leverage label information to disentangle inter-object features, thereby minimizing the interference of spurious features. Thus, the classification loss function is 
\begin{equation}
\mathcal{L}_{\mathrm{R/D}} = min \,\,\, \mathcal{L}_{\mathrm{CE}}(Y_{mab}^{\mathrm{RGB/Depth}}, Y),
\label{we1}
\end{equation}
where $\mathcal{L}_{\mathrm{CE}} (\cdot)$ denotes cross-entropy loss; $Y_{mab}^{\mathrm{RGB}}$ and $Y_{mab}^{\mathrm{Depth}}$ represent the outputs of the RGB and depth image classifiers, respectively; $Y$ is the ground-truth label.

\subsection{Information Bottleneck Fusion Module}
Once the MRN reconstructs the abnormal RGB and depth images synthesized by the MFEN into normal features using the additional features provided by the Mamba encoders, the cascade and cross-attention mechanisms are employed to fuse both modalities, as detailed follows:
\begin{equation}
\begin{split}
F_{fusion}^1=f_R^{rec2}\oplus f_D^{rec2}&,F_{fusion}^2=f_R^{rec4}\oplus f_D^{rec4},\\
F_{fu}=Cross\_&Att(F_{fusion}^1,F_{fusion}^2).
\end{split}
  \label{equ:21_111}
\end{equation}

Then, we use an information bottleneck regularization module to filter redundant information from the initial fusion feature (i.e., $F_{fu}$), obtaining a more predictive feature (i.e., $F_{fu}^g$). This module comprises two linear projection layers, dropout, and ReLU activation. Specifically, $z$ denotes the reprojection of $F_{fu}^g$ back to the dimension of $F_{fu}$, as:
\begin{equation}
F_{fu} \xrightarrow[\text{Linear Projection}]{\text{Dropout + ReLU}} z \xrightarrow[\text{Linear Projection}]{\text{Dropout + ReLU}} F_{fu} ^g.
\end{equation}

To ensure that $F_{fu}^g$ sufficiently retains the predictive information of $F_{fu}$ while discarding redundant features, we quantify this relationship using the mutual information between $F_{fu}^g$ and $F_{fu}$, defined as follows:
\begin{equation}
I(F_{fu};F_{fu}^g)=\mathbb{E}_{p(F_{fu},F_{fu}^g)}\left[\log\frac{p(F_{fu},F_{fu}^g)}{p(F_{fu})p(F_{fu}^g)}\right].
\label{eq7}
\end{equation}
Moreover, $I(F_{fu}; F_{fu}^g)$ can be futher divide into two parts, $I(F_{fu}; F_{fu}^g|Y)$ and $I( F_{fu}^g; Y)$, via the mutual information chain rule (See \textbf{\textit{Corollary} 1} in \textbf{Theoretical Analysis Section} for details), where $I(F_{fu}; F_{fu}^g|Y)$ represents redundant information and $I( F_{fu}^g; Y)$ indicates prediction-related information for the current object. Thus, to effectively eliminate redundant features, we only need to maximize $I( F_{fu}^g; Y)$ while minimizing $I(F_{fu}; F_{fu}^g|Y)$. Given $F_{fu}^g$  derived from $F_{fu}$, the information contained in $F_{fu}^g$ cannot exceed that in $F_{fu}$, i.e., $I(F_{fu}^g;Y) \leq I(F_{fu};Y)$. Consequently, the above information bottleneck objective function is equivalent to:
\begin{equation}
min\quad I(F_{fu};Y)-I(F_{fu}^g;Y). 
\label{eq1}
\end{equation}

\begin{table*}[t!]
    \centering
        \caption{I-AUROC/AUPRO results of our approach on MVTec 3D-AD under four incremental unified anomaly detection settings.}
    \renewcommand\arraystretch{1.2} 
    \setlength{\tabcolsep}{1.5mm}{
    \resizebox{\linewidth}{!}{
    	{\begin{tabular}{c|l|ccc|ccc|ccc|ccc}
    		\toprule[1.0pt]
    	\rowcolor{gray!10}&  &\multicolumn{3}{c|}{10-0 with 0 step (setting 1) }&\multicolumn{3}{c|}{9-1 with 1 step (setting 2)} &\multicolumn{3}{c|}{6-4 with 1 step (setting 3)}&\multicolumn{3}{c}{6-1 with 4 steps (setting 4)}\\ 
    \rowcolor{gray!10}&\multirow{-2}{*}{Method}&I-AUROC   &AUPRO &FM (\textbf{\textcolor{ForestGreen}{$\Downarrow$}})  &I-AUROC   &AUPRO &FM (\textbf{\textcolor{ForestGreen}{$\Downarrow$}}) &I-AUROC   &AUPRO &FM (\textbf{\textcolor{ForestGreen}{$\Downarrow$}})
        &I-AUROC    &AUPRO &FM (\textcolor{ForestGreen}{$\Downarrow$})\\ 
            \midrule[0.5pt]
            \multirow{3}{*} 
            {\rotatebox{90}{RGB}}
            &IUF \scriptsize (ECCV24) &87.1 &88.5 &-- / -- &82.4&85.1 & 2.9 / 1.5 &76.7&78.6 &8.3 / 6.9 &73.9 &76.3&12.5 / 7.9\\
            &CDAD \scriptsize (CVPR25) &76.4 &87.5 &-- / --&74.2 &83.9&2.1 / 1.9 &68.9&77.5 &7.4 / 6.3 &67.4 &73.9 &\secondcolor{\textbf{\textcolor{ForestGreen}{8.0}} / 6.7}\\
            &IB-IUMAD &\firstcolor{\textbf{88.9}}  &\firstcolor{\textbf{89.3}} &\secondcolor{-- / --}
            &\firstcolor{\textbf{85.3}} &\firstcolor{\textbf{87.6}} &\secondcolor{\textbf{\textcolor{ForestGreen}{1.8}} / \textbf{\textcolor{ForestGreen}{1.3}} } &\firstcolor{\textbf{79.8}}
            &\firstcolor{\textbf{81.2}} &\textbf{\textcolor{ForestGreen}{6.5}} / \secondcolor{\textbf{\textcolor{ForestGreen}{5.8}}} &\firstcolor{\textbf{75.7}} &\firstcolor{\textbf{80.5}}
            &8.7 / \secondcolor{\textbf{\textcolor{ForestGreen}{6.3}}}\\
            \midrule[0.5pt]
            \multirow{3}{*}
            {\rotatebox{90}{3D}}
            &IUF \scriptsize (ECCV24) &73.7 &65.4 &-- / -- &70.4 &62.1 &1.6 / 1.4&67.4&59.2 &4.4 / 3.6 &66.8 &59.7 &4.7 / 3.0\\
            &CDAD \scriptsize (CVPR25) &60.6 &40.6 &-- / --&58.8 &39.2 &\secondcolor{1.1 / \textbf{\textcolor{ForestGreen}{0.2}} }&55.5&37.9 &1.7 / 0.9&56.1 &38.5 &2.5 / \textbf{1.4}\\
            &IB-IUMAD &\firstcolor{\textbf{75.8}} &\firstcolor{\textbf{67.6}} &\secondcolor{-- / --}
            &\firstcolor{\textbf{74.9}} &\firstcolor{\textbf{65.5}} &\secondcolor{\textbf{\textcolor{ForestGreen}{-0.3}} / 0.7 }&\firstcolor{\textbf{72.6}}
            &\firstcolor{\textbf{64.3}} &\secondcolor{\textbf{\textcolor{ForestGreen}{1.3}} / \textbf{\textcolor{ForestGreen}{0.2}}}  &\firstcolor{\textbf{71.2}} &\firstcolor{\textbf{64.1}} &\secondcolor{\textbf{\textcolor{ForestGreen}{2.2}} / \textbf{\textcolor{ForestGreen}{1.4}}}\\
            \midrule[0.5pt]
            \multirow{3}{*}
            {\rotatebox{90}{RGB+3D}}
            &IUF \scriptsize (ECCV24) &88.7 &89.2 &-- / --&84.2 &86.6 &3.7 / 2.2 &79.2&80.2 &10.6 / 7.3 &75.1 &79.5 &15.1 / 8.4\\
            &CDAD \scriptsize (CVPR25) &79.1 &88.1 &-- / --&75.1 &85.3 &3.4 / 2.1 &70.1&78.5 &8.5 / 7.1&69.5 &75.7 &\secondcolor{\textbf{\textcolor{ForestGreen}{8.9}} / 7.7}\\
            &IB-IUMAD &\firstcolor{\textbf{91.0}} &\firstcolor{\textbf{90.4}} &\secondcolor{-- / --}
            &\firstcolor{\textbf{87.5}} &\firstcolor{\textbf{89.1}} &\secondcolor{\textbf{\textcolor{ForestGreen}{2.3}} / \textbf{\textcolor{ForestGreen}{0.8}}} &\secondcolor{\firstcolor{\textbf{82.4}}}
             &\firstcolor{\textbf{84.7}}
            &\secondcolor{\textbf{\textcolor{ForestGreen}{6.8}} / \textbf{\textcolor{ForestGreen}{6.4}}} &\firstcolor{\textbf{78.6} }&\firstcolor{\textbf{82.4}} &\secondcolor{9.3 / \textbf{\textcolor{ForestGreen}{6.9}}}\\
            \bottomrule[1.0pt]
    \end{tabular}}}}
\end{table*}
\begin{table*}[h]
    \caption{I-AUROC/AUPRO results of our approach on Eyecandies under four incremental unified anomaly detection settings.}
    \centering
    \renewcommand\arraystretch{1.2} 
    \setlength{\tabcolsep}{1.5mm}{
    \resizebox{\linewidth}{!}{
    	{\begin{tabular}{c|l|ccc|ccc|ccc|ccc}
    		\toprule[1.0pt]
    	\rowcolor{gray!10}&  &\multicolumn{3}{c|}{10-0 with 0 step (setting 1)} &\multicolumn{3}{c|}{9-1 with 1 step (setting 2)} &\multicolumn{3}{c|}{6-4 with 1 step (setting 3)}&\multicolumn{3}{c}{6-1 with 4 steps (setting 4)}\\ 
\rowcolor{gray!10}&\multirow{-2}{*}{Method}&I-AUROC   &AUPRO &FM (\textbf{\textcolor{ForestGreen}{$\Downarrow$}}) &I-AUROC    &AUPRO &FM (\textbf{\textcolor{ForestGreen}{$\Downarrow$}}) &I-AUROC    &AUPRO &FM (\textbf{\textcolor{ForestGreen}{$\Downarrow$}})
        &I-AUROC    &AUPRO &FM (\textbf{\textcolor{ForestGreen}{$\Downarrow$}})\\ 
            \midrule[0.5pt]
            \multirow{3}{*}{\rotatebox{90}{RGB}}
            &IUF \scriptsize (ECCV24) &77.2 &84.0 &-- / --&73.9 &79.6 &3.8 / 3.5 &62.9 &68.6&12.4 / 9.1 &63.6 &69.1 &10.6 / 8.8\\
            &CDAD \scriptsize (CVPR25) &78.1 &82.1 &-- / --&73.2 &76.5 &4.2 / 3.1 &63.7 &70.4 &9.5 / 7.7 &65.5&72.3 &8.4 / 6.3\\
            &IB-IUMAD &\firstcolor{\textbf{78.5}}  &\firstcolor{\textbf{85.3}} &\secondcolor{-- / --}
            &\firstcolor{\textbf{74.1}}
            &\firstcolor{\textbf{83.2}} &\secondcolor{\textbf{\textcolor{ForestGreen}{2.7}} / \textbf{\textcolor{ForestGreen}{1.1}}} &\firstcolor{\textbf{65.4}} &\firstcolor{\textbf{71.2}}
            &\secondcolor{\textbf{\textcolor{ForestGreen}{8.1}} / \textbf{\textcolor{ForestGreen}{6.5}}} &\firstcolor{\textbf{66.5}} &\firstcolor{\textbf{72.7}} &\secondcolor{\textbf{\textcolor{ForestGreen}{7.3}} / \textbf{\textcolor{ForestGreen}{5.7}}}\\
            \midrule[0.5pt]
            \multirow{3}{*}{\rotatebox{90}{3D}}
            &IUF \scriptsize (ECCV24) &57.2  &39.4 &-- / --&55.6 &36.4 &2.1 / 0.8 &51.5&37.4 &3.8 / 1.2 &51.9 &38.2 &3.0 / 1.1\\
            &CDAD \scriptsize (CVPR25) &58.4  &30.2 &-- / --&57.1 &29.3 &\secondcolor{\textbf{\textcolor{ForestGreen}{0.9}} / \textbf{\textcolor{ForestGreen}{0.1}}} &\firstcolor{\textbf{53.1}}&29.6 &\secondcolor{4.2 / \textbf{\textcolor{ForestGreen}{-0.4}}} &52.9  &30.1 &\secondcolor{4.4 / \textbf{\textcolor{ForestGreen}{0.3}}}\\
            &IB-IUMAD &\firstcolor{\textbf{59.2}}  &\firstcolor{\textbf{43.7}} &\secondcolor{-- / --}
             &\firstcolor{\textbf{57.5}}
            &\firstcolor{\textbf{39.8}} &1.4 / 1.3 &52.9 &\firstcolor{\textbf{41.3}}
            &\secondcolor{\textbf{\textcolor{ForestGreen}{3.5}} / 1.4} &\firstcolor{\textbf{53.3}} &\firstcolor{\textbf{41.9}} &\secondcolor{\textbf{\textcolor{ForestGreen}{2.6}} / 0.9}\\
            \midrule[0.5pt]
            \multirow{3}{*}{\rotatebox{90}{RGB+3D}}
            &IUF \scriptsize (ECCV24) &78.4  &85.5 &-- / --&74.4 &80.3 &4.3 / 3.8 &64.1 &69.8&14.3 / 8.6 &65.8 &71.5 &12.5 / 8.8\\
            &CDAD \scriptsize (CVPR25) &79.2 &83.7&-- / --&74.1 &78.8 &4.6 / 3.5 &66.5 &72.3&10.1 / 7.9 &68.2 &73.8 &9.2 / 6.6\\
            &IB-IUMAD &\firstcolor{\textbf{80.6}}  &\firstcolor{\textbf{86.1}} &\secondcolor{-- / --}
            &\firstcolor{\textbf{76.7}} 
            &\firstcolor{\textbf{85.6}} &\secondcolor{\textbf{\textcolor{ForestGreen}{3.2}} / \textbf{\textcolor{ForestGreen}{1.6}}} &\firstcolor{\textbf{67.8}} &\firstcolor{\textbf{73.1}}
            &\secondcolor{\textbf{\textcolor{ForestGreen}{8.4}} / \textbf{\textcolor{ForestGreen}{6.7}}} &\firstcolor{\textbf{69.4}} &\firstcolor{\textbf{75.1}} &\secondcolor{\textbf{\textcolor{ForestGreen}{7.7}} / \textbf{\textcolor{ForestGreen}{6.3}}}\\
            \bottomrule[1.0pt]
    \end{tabular}}}}
\end{table*}
Finally, to optimize the above objective function Eq. \ref{eq1}, we adopt Kullback-Leibler divergence as the loss function to optimize feature $F_{fu}^g$ (See \textbf{\textit{Corollary} 2} in \textbf{Theoretical Analysis Section} for details), defined as:
\begin{equation}
\mathcal{L}_{\mathrm{IB}} = \mathrm{KL} [P(Y|F_{fu}) || P(Y|F_{fu}^g)].
\end{equation}

In summary, for the IUMAD task, we first employ the classic MSE loss for reconstruction, followed by the cross-entropy loss for classification on both RGB and depth images to prevent interference from spurious features. To further mitigate redundant information in the fused features, we use the KL divergence loss. Thus, the overall loss can be formulated as:
\begin{equation}
\begin{split}
&\mathcal{L}_{\mathrm{Fusion}} = \frac{1}{W \times H} \left\|F_{\mathrm{org}}^{\mathrm{RGB}} - F_{fusion}^g\right\|_2^2,
\end{split}
\label{equ111}
\end{equation}
\begin{equation}
\begin{split}
\mathcal{L}_{\mathrm{All}} &= \lambda_1\mathcal{L}_{\mathrm{CE}}(Y_{mab}^{\mathrm{RGB}}, Y) + \lambda_2\mathcal{L}_{\mathrm{CE}}(Y_{mab}^{\mathrm{Depth}}, Y) \\
&\phantom{=} \underbrace{+\ \lambda_3\mathcal{L}_{\mathrm{Fusion}} + \lambda_4\mathcal{L}_{\mathrm{IB}}(Y_{F_{fu}}, Y_{F_{fu}^g})}_{\mathrm{Information\ Bottleneck\ Fusion\ Module}},
\end{split}
\label{equ111222}
\end{equation}
where $\lambda_1$, $\lambda_2$, $\lambda_3$, and $\lambda_4$ are loss balance hyperparameters, respectively, and all are set to 1 in our experiments.

\subsection{Theoretical Analysis}
Further, we theoretically analyze the effectiveness of the information bottleneck regularization in filtering redundant information within the multimodal fusion feature. For this purpose, we follow the information theory \cite{iclr/Federici20,tian2021farewell,fang2024dynamic} and prepare several corollaries. 

\begin{figure}[h]
    \begin{tcolorbox}[title=Corollary 1.]
\noindent Given a random variable $Y$, the mutual information between features $F_{fu}$ and $F_{fu}^g$, expressed in Eq. \ref{eq7},  can be equivalently rewritten as $I(F_{fu}; F_{fu}^g)=$ $I(F_{fu}; F_{fu}^g|Y) +I(F_{fu}^g;Y)$ based on the chain rule.\\
        
        \noindent\emph{\textbf{Proof}.} Due to the limitation of pages, the detailed proof is provided in the Appendix \textcolor{blue}{A}.
    \end{tcolorbox}
\end{figure}
\vspace{-10pt}
\begin{figure}[!h]
    \begin{tcolorbox}[title=Corollary 2.]
        \noindent If the KL divergence between $p(Y|F_{fu})$ and $p(Y|F_{fu}^g)$ is 0, that is, $\mathrm{KL} [P(Y|F_{fu})||$ $ P(Y|F_{fu}^g)]=0$, we have $I(Y;F_{fu})-I(Y;F_{fu}^g) =I(F_{fu};Y)-I(F_{fu}^g;Y) =0$.
        \\
        
        \emph{\textbf{Proof}.} $I(Y;F_{fu})-I(Y;F_{fu}^g)$
\vspace{-10pt}
\begin{align*}
&=\iint P(F_{fu}) P(Y|F_{fu}) \log P(Y|F_{fu}) d F_{fu}dY \\
&-\iint P(F_{fu}^g)P(Y|F_{fu}^g) \log P(Y|F_{fu}^g) d F_{fu}^g dY\\
&=\mathbb{E}_{F_{fu}}[\mathrm{KL}[P(Y|F_{fu})||P(Y|F_{fu}^g)]] -\mathbb{E}_{F_{fu}^g}[\mathrm{KL}[P\\
&(Y|F_{fu}^g)||P(Y|F_{fu})]] +\int P(Y)\log \frac{P(Y|F_{fu}^g)}{p(Y|F_{fu})} dY.
\end{align*}
\end{tcolorbox}
\end{figure}


By Jensen’s inequality and the strict convexity of $-\log$, we conclude that the KL divergence is non-negative. When $\mathrm{KL} [P(Y|F_{fu}) || P(Y|F_{fu}^g)]=0$, it follows that $P(Y|F_{fu}^g)$ $=P(Y|F_{fu})$ almost everywhere, which means that $\int P(Y) \log \frac{P(Y|F_{fu}^g)}{p(Y|F_{fu})} dY =0$, and we have $I(Y|F_{fu})-I(Y|F_{fu}^g)= 0$. Therefore, based on \textbf{\textit{Corollary} 1} and \textbf{\textit{Corollary} 2}, we conclude that using KL divergence as the target loss function between $F_{fu}$ and $F_{fu}^g$ effectively eliminates redundant information from the multimodal fusion feature.

\section{Evaluation}
\label{sec:experi}
We then conduct empirical comparisons to validate the effectiveness of IB-IUMAD.
\subsection{Dataset and Experimental Setting}
\textbf{Dataset.}\quad We evaluate IB-IUMAD on MVTec 3D-AD \cite{DBLP:conf/visapp/BergmannJSS22} and Eyecandies \cite{bonfiglioli2022eyecandies} datasets. Both contain 10 objects, where the former is collected from realistic scenes, and the latter is synthetic.

\textbf{Setup.}\quad Firstly, we conducted comparison experiments of IB-IUMAD with incremental unified methods such as IUF \cite{tang2024incremental} and CDAD \cite{li2025one} in four different settings: 10-0 with 0 step, 9-1 with 1 step, 6-4 with 1 step, and 6-1 with 4 steps. Then, for comparison with unified MAD methods, we report the results of IB-IUMAD (i.e., 10-0 with 0 step setting) with UniAD \cite{you2022unified}, SimpleNet \cite{liu2023simplenet}, DeSTSeg \cite{zhang2023destseg}, DiAD \cite{he2024diffusion}, and MambaAD \cite{DBLP:conf/nips/HeBZHCGWLT024}. For implementation details, please refer to Appendix \textcolor{blue}{F}.

\begin{table*}[ht!]
\centering
\caption{I-AUROC scores on  MVTec 3D-AD dataset (10-0 with 0 step). The \colorbox{firstcolor}{red}/\colorbox{secondcolor}{blue} indicates the best/second-best results.}
\setlength{\tabcolsep}{2.8mm} 
\renewcommand{\arraystretch}{0.9}
\adjustbox{width=1\linewidth}{
   \begin{tabular}{c|l|c|c ccc ccc ccc |c}
       \toprule
\rowcolor{gray!10}& &&  &  &  &  &  &  &  &   &  & & \\
\rowcolor{gray!10}\multirow{-2}{*}{}
                 & \multirow{-2}{*}{Method}
                 & \multirow{-2}{*}{Year}
                &\multirow{-2}{*}{Bagel}
                &\multirow{-2}{*}{Cable Gland} 
                &\multirow{-2}{*}{Carrot} 
                &\multirow{-2}{*}{Cookie} 
                &\multirow{-2}{*}{Dowel} 
                &\multirow{-2}{*}{Foam} 
                &\multirow{-2}{*}{Peach} 
                &\multirow{-2}{*}{Potato} 
                &\multirow{-2}{*}{Rope} 
                &\multirow{-2}{*}{Tire}
                 &\multirow{-2}{*}{Mean} \\
\midrule[0.5pt]  
\multirow{8}{*}{\rotatebox{90}{RGB}}     
 &UniAD  &NIPS22 &82.7 &89.8 &76.8 &77.3 &96.7 &70.5 &70.0 &51.6 &\firstcolor{\textbf{97.4}} &75.7 &78.9\\
&SimpleNet  &CVPR23 &76.2 &70.3 &71.4 &66.7 &83.7 &77.4 &62.0 &56.7 &95.6 &65.3 &72.5\\
&DeSTSeg  &CVPR23 &89.7 &84.8 &79.1 &67.4 &77.3 &77.9 &82.2 &62.9 &93.5 &80.8 &79.6\\
&MambaAD  &NIPS24 &87.7 &\secondcolor{\textbf{\textcolor{ForestGreen}{94.3}}} &90.7 &61.2 &\secondcolor{\textbf{\textcolor{ForestGreen}{97.6}}} &\secondcolor{\textbf{\textcolor{ForestGreen}{84.0}}} &\secondcolor{\textbf{\textcolor{ForestGreen}{92.8}}} &66.8 &\firstcolor{\textbf{97.4}} &\secondcolor{\textbf{\textcolor{ForestGreen}{90.0}}} &86.2\\
&DiAD  &AAAI24 &\firstcolor{\textbf{100.0}} &68.1 &\secondcolor{\textbf{\textcolor{ForestGreen}{94.4}}} &69.4 &\firstcolor{\textbf{98.0}} &\firstcolor{\textbf{100.0}} &58.0 &76.3 &89.2 &\firstcolor{\textbf{92.7}} &84.6\\
&IUF  &ECCV24 &88.3 &\firstcolor{\textbf{95.0}} &\firstcolor{\textbf{100.0}} &\firstcolor{\textbf{89.7}} &93.7 &73.6 &83.3 &\firstcolor{\textbf{93.1}} &91.1 &63.5 &\secondcolor{\textbf{\textcolor{ForestGreen}{87.1}}}\\
&CDAD  &CVPR25 &87.4 &88.4 &74.9 &67.6 &91.2 &70.7 &64.2 &59.9 &86.2 &73.8 &76.4\\
& IB-IUMAD &- & \secondcolor{\textbf{\textcolor{ForestGreen}{95.4}}} & 93.1 & 90.0& \secondcolor{\textbf{\textcolor{ForestGreen}{82.1}}}& 92.9& 68.4& \firstcolor{\textbf{94.6}}& \secondcolor{\textbf{\textcolor{ForestGreen}{90.4}}}& \secondcolor{\textbf{\textcolor{ForestGreen}{96.7}}}& 85.6 & \firstcolor{\textbf{88.9}}\\
\midrule[0.5pt]
\multirow{4}{*}{\rotatebox{90}{3D}}
            &DiAD  &AAAI24 &\firstcolor{\textbf{79.4}} &59.2 &\firstcolor{\textbf{71.7}} &65.2 &\firstcolor{\textbf{78.4}} &74.5 &\secondcolor{\textbf{\textcolor{ForestGreen}{56.3}}} &65.0 &60.1 &\secondcolor{\textbf{\textcolor{ForestGreen}{64.2}}} &67.4\\
            &IUF  &ECCV24 &68.9 &\secondcolor{\textbf{\textcolor{ForestGreen}{96.5}}} &\secondcolor{\textbf{\textcolor{ForestGreen}{68.4}}} &\secondcolor{\textbf{\textcolor{ForestGreen}{73.5}}} &\secondcolor{\textbf{\textcolor{ForestGreen}{75.4}}} &\secondcolor{\textbf{\textcolor{ForestGreen}{76.7}}} &55.0 &\secondcolor{\textbf{\textcolor{ForestGreen}{72.3}}} &\firstcolor{\textbf{89.1}} &61.4 &\secondcolor{\textbf{\textcolor{ForestGreen}{73.7}}}\\
            &CDAD  &CVPR25  &64.7 &65.2 &54.7 &68.9 &56.3 &62.7 &55.1 &51.9 &58.6&\firstcolor{\textbf{67.7}} &60.6\\

              &IB-IUMAD &- & \secondcolor{\textbf{\textcolor{ForestGreen}{69.1}}} & \firstcolor{\textbf{99.3}} & 63.6& \firstcolor{\textbf{74.8}}& 65.5& \firstcolor{\textbf{82.2}}& \firstcolor{\textbf{69.7}}& \firstcolor{\textbf{96.4}}& \secondcolor{\textbf{\textcolor{ForestGreen}{76.6}}}& 60.4 & \firstcolor{\textbf{75.8}}\\
\midrule[0.5pt]
\multirow{4}{*}{\rotatebox{90}{RGB + 3D}}
            &DiAD  &AAAI24 &\firstcolor{\textbf{100.0}} &73.9 &\firstcolor{\textbf{97.2}} &71.6 &\firstcolor{\textbf{97.6}} &\firstcolor{\textbf{98.7}} &69.4  &\secondcolor{\textbf{\textcolor{ForestGreen}{78.3}}} &\secondcolor{\textbf{\textcolor{ForestGreen}{94.3}}} &\secondcolor{\textbf{\textcolor{ForestGreen}{85.6}}} &86.7\\
            &IUF  &ECCV24 &97.2 &\secondcolor{\textbf{\textcolor{ForestGreen}{96.9}}} &\secondcolor{\textbf{\textcolor{ForestGreen}{89.1}}} &\secondcolor{\textbf{\textcolor{ForestGreen}{75.4}}} &\secondcolor{\textbf{\textcolor{ForestGreen}{93.4}}} &\secondcolor{\textbf{\textcolor{ForestGreen}{89.7}}} &\secondcolor{\textbf{\textcolor{ForestGreen}{95.6}}} &74.2 &\firstcolor{\textbf{98.4}} &77.3 &\secondcolor{\textbf{\textcolor{ForestGreen}{88.7}}}\\
            &CDAD  &CVPR25 &90.3 &89.7 &78.2 &68.4 &92.5 &73.4 &65.9 &63.3 &89.4 &79.7 &79.1\\
            & IB-IUMAD &- &\secondcolor{\textbf{\textcolor{ForestGreen}{98.3}}} & \firstcolor{\textbf{97.5}} & 86.3& \firstcolor{\textbf{84.6}}& 90.5& 69.4& \firstcolor{\textbf{98.7}}& \firstcolor{\textbf{100.0}}& \firstcolor{\textbf{98.4}}& \firstcolor{\textbf{86.4}} & \firstcolor{\textbf{91.0}}\\
             \bottomrule
       \end{tabular}
   }
\end{table*}
\begin{table*}[ht!]
\centering
   \caption{I-AUROC scores on  Eyecandies dataset (10-0 with 0 step). The \colorbox{firstcolor}{red}/\colorbox{secondcolor}{blue} indicates the best/second-best results.}
\setlength{\tabcolsep}{2.0mm} 
\renewcommand{\arraystretch}{0.9}
\adjustbox{width=1\linewidth}{
   \begin{tabular}{c|l|c|c ccc ccc ccc |c}
       \toprule
                 \rowcolor{gray!10}\multirow{2}{*}{}
                 & 
                 &
                &\multirow{1}{*}{Candy}
                &\multirow{1}{*}{Chocolate} 
                &\multirow{1}{*}{Chocolate} 
                & 
                &\multirow{1}{*}{Gummy} 
                &\multirow{1}{*}{Hazelnut} 
                &\multirow{1}{*}{Licorice} 
                &
                &\multirow{1}{*}{Marsh-} 
                &\multirow{1}{*}{Peppermint}
                 & \\
                \rowcolor{gray!10} &\multirow{-2}{*}{Method} & \multirow{-2}{*}{Year}&  Cane  &Cookie  &Praline  &\multirow{-2}{*}{Confetto}  &Bear  &Truffle  &Sandwish  &\multirow{-2}{*}{Lollipop}    &mallow  &Candy &\multirow{-2}{*}{Mean}\\
\midrule[0.5pt]  
\multirow{4}{*}{\rotatebox{90}{RGB}}     
            &DiAD  &AAAI24 &56.6 &83.5 &77.3 &\secondcolor{\textbf{\textcolor{ForestGreen}{89.7}}} &59.8 &\secondcolor{\textbf{\textcolor{ForestGreen}{59.4}}} &\secondcolor{\textbf{\textcolor{ForestGreen}{84.3}}} &66.7 &97.4 &86.8 &76.2\\
            &IUF  &ECCV24 &56.3 &86.2 &\secondcolor{\textbf{\textcolor{ForestGreen}{78.7}}} &\firstcolor{\textbf{90.8}} &59.5 &\firstcolor{\textbf{61.0}} &84.0 &\secondcolor{\textbf{\textcolor{ForestGreen}{68.8}}} &\secondcolor{\textbf{\textcolor{ForestGreen}{98.1}}} &\secondcolor{\textbf{\textcolor{ForestGreen}{88.3}}} &77.2\\
            &CDAD  &CVPR25 &\secondcolor{\textbf{\textcolor{ForestGreen}{62.2}}} &\firstcolor{\textbf{92.1}} &78.2 &87.4 &\firstcolor{\textbf{68.3}} &56.1 &\firstcolor{\textbf{85.4}} &\firstcolor{\textbf{70.6}} &95.8 &85.2 &\secondcolor{\textbf{\textcolor{ForestGreen}{78.1}}}\\
              & IB-IUMAD &- & \firstcolor{\textbf{63.7}} & \secondcolor{\textbf{\textcolor{ForestGreen}{88.6}}} & \firstcolor{\textbf{82.4}}& 87.8& \secondcolor{\textbf{\textcolor{ForestGreen}{64.4}}}& 56.9& 82.7& 67.7& \firstcolor{\textbf{98.7}}& \firstcolor{\textbf{92.4}} & \firstcolor{\textbf{78.5}}\\
\midrule[0.5pt]
\multirow{4}{*}{\rotatebox{90}{3D}}
            &DiAD  &AAAI24 &51.8 &57.1 &55.0 &58.1 &56.6 &50.4 &52.9 &\secondcolor{\textbf{\textcolor{ForestGreen}{56.8}}} &54.9 &54.1 &54.8\\
            &IUF  &ECCV24 &\secondcolor{\textbf{\textcolor{ForestGreen}{52.5}}} &53.1 &\firstcolor{\textbf{65.8}} &55.8 &\firstcolor{\textbf{58.6}} &\firstcolor{\textbf{65.8}} &52.2 &53.9 &53.9 &\secondcolor{\textbf{\textcolor{ForestGreen}{63.7}}} &57.5\\
            &CDAD  &CVPR25 &\firstcolor{\textbf{55.9}} &\firstcolor{\textbf{62.8}} &50.9 &\firstcolor{\textbf{65.9} }&52.2&54.4 &\firstcolor{\textbf{61.9}} &\firstcolor{\textbf{66.7}} &\secondcolor{\textbf{\textcolor{ForestGreen}{55.4}}} &58.3&\secondcolor{\textbf{58.4}}\\

              & IB-IUMAD &- & 51.2 & \secondcolor{\textbf{\textcolor{ForestGreen}{57.4}}} & \secondcolor{\textbf{\textcolor{ForestGreen}{65.6}}}& \secondcolor{\textbf{\textcolor{ForestGreen}{59.8}}}& \secondcolor{\textbf{\textcolor{ForestGreen}{57.7}}}&\secondcolor{\textbf{\textcolor{ForestGreen}{65.6}}}& \secondcolor{\textbf{\textcolor{ForestGreen}{53.2}}}& 56.1& \firstcolor{\textbf{60.9}}& \firstcolor{\textbf{64.7}} & \firstcolor{\textbf{59.2}}\\
\midrule[0.5pt]
\multirow{4}{*}{\rotatebox{90}{RGB + 3D}}
            &DiAD  &AAAI24 &54.1 &87.5 &78.2 &\firstcolor{\textbf{91.7}} &58.2 &55.7 &\secondcolor{\textbf{\textcolor{ForestGreen}{85.0}}} &71.9 &\secondcolor{\textbf{\textcolor{ForestGreen}{98.8}}} &88.3 &76.9\\
            &IUF  &ECCV24 &\firstcolor{\textbf{67.3}} &87.2 &\secondcolor{\textbf{\textcolor{ForestGreen}{81.6}}} &85.6 &61.7 &\secondcolor{\textbf{\textcolor{ForestGreen}{57.9}}} &81.6 &66.3 &98.2 &\firstcolor{\textbf{96.3} }&78.4\\
            &CDAD  &CVPR25 &\secondcolor{\textbf{\textcolor{ForestGreen}{66.9}}} &\firstcolor{\textbf{91.4}} &75.3 &\secondcolor{\textbf{\textcolor{ForestGreen}{88.9}}} &\firstcolor{\textbf{74.2}} &55.8 &\firstcolor{\textbf{86.1}} &\firstcolor{\textbf{77.1}} &94.0 &82.5 &\secondcolor{\textbf{\textcolor{ForestGreen}{79.2}}}\\
            & IB-IUMAD &- & 66.5 & \secondcolor{\textbf{\textcolor{ForestGreen}{90.1}}} & \firstcolor{\textbf{84.6}}& 87.7& \secondcolor{\textbf{\textcolor{ForestGreen}{62.3}}}& \firstcolor{\textbf{59.7}}&84.2& \secondcolor{\textbf{\textcolor{ForestGreen}{75.4}}}& \firstcolor{\textbf{99.4}}& \secondcolor{\textbf{\textcolor{ForestGreen}{95.6}}} & \firstcolor{\textbf{80.6}}\\
             \bottomrule
       \end{tabular}
   }
\end{table*}
\subsection{Quantitative Evaluation}
\textbf{Comparison of Incremental Settings.}\quad Tables 1 and 2 present performance comparisons between IB-IUMAD and other incremental unified methods on the MVTec 3D-AD and Eyecandies datasets, respectively. The results show that IB-IUMAD consistently outperforms the baselines across different methods, incremental settings, and datasets. Specifically, when trained with RGB images on the MVTec 3D-AD under setting 2, IB-IUMAD surpasses CDAD by 11.1\% and 3.7\% in I-AUROC and AUPRO, respectively, while significantly dropping the FM metric by 0.3\% and 0.6\%. When trained with RGB and depth images on the MVTec 3D-AD (Eyecandies) under setting 4, IB-IUMAD outperforms IUF by 3.5\% and 2.9\% (3.6\% and 3.6\%) in I-AUROC and AUPRO, respectively, and significantly reduces the FM metric by 5.8\% and 1.5\% (4.8\% and 2.5\%). These results highlight the effectiveness of our devised denoising module in mitigating the impact of spurious and redundant features on model performance and catastrophic forgetting.

\textbf{Comparison of Unified MAD.}\quad To further demonstrate the effectiveness of the proposed approach, Tables 3 and 4 present the experimental results of IB-IUMAD (setting 1) in comparison with the unified MAD baselines across both datasets. As shown, IB-IUMAD achieves superior detection results in most test cases under unimodal and multimodal settings. Notably, when both RGB and depth images are utilized for training, IB-IUMAD improves the I-AUROC by 4.3\% and 2.3\% on MVTec 3D-AD, and by 3.7\% and 2.2\% on Eyecandies, compared to DiAD and IUF, respectively. These results, along with the visualizations in Figure 4, further validate the effectiveness of our proposed approach in improving unified MAD. Due to page limitations, more experimental results are provided in Appendices \textcolor{blue}{B} and \textcolor{blue}{C}\footnote{Appendix released at \href{https://github.com/longkaifang/IB-IUMAD}{https://github.com/longkaifang/IB-IUMAD}}.

\subsection{Ablation Study}
\textbf{Impact of Key Components.}\quad In this part, we scrutinize the contribution of each component within IB-IUMAD to incremental unified tasks. As shown in Table 5, IB-IUMAD obtains consistently better results when both the Mamba decoder and IBFM module are used, compared to using only one of them or neither. For example, on 6-1 with 4 steps, I-AUROC increases by 2.5\% and FM decreases by 1.9\%; on 9-1 with 1 step, I-AUROC improves by 2.4\% and FM decreases by 2.3\%, compared to the variant without both Mamba and IBFM. It is clear that our proposed method is extremely competitive in incremental unified framework.

\textbf{Impact of Fusion Operations.}\quad To reveal the impact of various fusion operations in the IBFM module on model performance, we conducted several groups of tests with different fusion operations on the MVTec 3D-AD dataset with setting 1. As shown in Table 6, the cross-attention mechanism achieves better results with 91.0\%, 97.6\%, and 90.4\% on the I-AUROC, P-AUROC, and AUPRO metrics, respectively. This further shows that well-designed multimodal fusion operations equipped with information bottleneck constraints can effectively improve the IUMAD's overall performance.

\textbf{Computing Efficiency.}\quad To further show the advantages of IB-IUMAD, we report the results in terms of memory usage, frame rate and I-AUROC on MVTec 3D-AD. As shown in Table 7, compared to M3DM, IB-IUMAD gained a 41x increase in frame rate and 44x shrinkage in memory usage while maintaining comparable performance. This once again reveals the enormous benefits of our method in IUMAD tasks.
\begin{table}[t]
    \centering
    \caption{The ablation study results on the MVTec 3D-AD dataset.}
    \renewcommand\arraystretch{1.0}
    \setlength{\tabcolsep}{0.9mm}
    \resizebox{\linewidth}{!}{
        \begin{tabular}{cc|cc|cc|cc|c}
            \toprule[1.0pt]
            \rowcolor{gray!10}\multicolumn{2}{c|}{Ablation} & \multicolumn{2}{c|}{6-1 with 4} & \multicolumn{2}{c|}{6-4 with 1} & \multicolumn{2}{c|}{9-1 with 1} & \multicolumn{1}{c}{10-0 with 0} \\ 
            \rowcolor{gray!10}Mamba & IBFM &I-AUROC &FM &I-AUROC &FM &I-AUROC &FM & I-AUROC \\ 
            \midrule[0.5pt]
            \textcolor{ForestGreen}{\XSolidBrush} & \textcolor{ForestGreen}{\XSolidBrush} & 75.3 & 12.7 & 79.9 & 8.7 & 85.1 & 4.6 & 86.7 \\
            \textcolor{red}{\textbf{\Checkmark}} & \textcolor{ForestGreen}{\XSolidBrush} & 76.0 & 11.0 & 80.8 & 7.6 & 86.0 & 4.0 & 88.8 \\
            \textcolor{ForestGreen}{\XSolidBrush} & \textcolor{red}{\Checkmark} & 76.9 & 10.2 & 81.5 & 7.1 & 86.8 & 3.2 & 89.2 \\ 
             \textcolor{red}{\Checkmark} & \textcolor{red}{\Checkmark} & \firstcolor{\textbf{78.6}} & \secondcolor{\textbf{\textcolor{ForestGreen}{9.3}}} & \firstcolor{\textbf{82.4}} & \secondcolor{\textbf{\textcolor{ForestGreen}{6.8}}} & \firstcolor{\textbf{87.5}} & \secondcolor{\textbf{\textcolor{ForestGreen}{2.3}}} & \firstcolor{\textbf{91.0}} \\
            \bottomrule[1.0pt]
        \end{tabular}
    }
\end{table}

\setlength{\fboxrule}{1.3pt}
\begin{table}[t]
    \centering
    \caption{The impact of multimodal fusion operations.}
    \renewcommand\arraystretch{0.9} 
    \setlength{\tabcolsep}{5.5mm}{
    \resizebox{\linewidth}{!}{
    	{\begin{tabular}{l|c|c|c}
    		\toprule[1.0pt]
    	\rowcolor{gray!10}  &\multicolumn{3}{c}{MVTec 3D-AD} \\ 
    		\rowcolor{gray!10}\multirow{-2}{*}{Fusion types}&I-AUROC &P-AUROC &AUPRO \\ 
                \midrule[0.5pt]
Addition &\fcolorbox{secondcolor}{white}{89.8} 	&\fcolorbox{secondcolor}{white}{96.7}  &88.6  \\ 
ConcatFC &88.2 	&96.4  &\fcolorbox{secondcolor}{white}{89.5}  \\ 
LinearGLU &86.7 	&95.8  &88.3 \\  Cross-attention &\fcolorbox{firstcolor}{white}{\textbf{91.0}}	&\fcolorbox{firstcolor}{white}{\textbf{97.6}}	&\fcolorbox{firstcolor}{white}{\textbf{90.4}}\\
            \bottomrule[1.0pt]
    \end{tabular}}}}
\end{table}

\setlength{\fboxrule}{1.3pt}
\begin{table}[t!]
    \centering
    \caption{Comparison results in terms of the computing efficiency.}
    \renewcommand\arraystretch{0.9} 
    \setlength{\tabcolsep}{6.0mm}{
    \resizebox{\linewidth}{!}{
    	{\begin{tabular}{l|c|c|cc}
    		\toprule[1.0pt]
    	\rowcolor{gray!20}  &\multicolumn{3}{c}{MVTec 3D-AD} \\ 
    	\rowcolor{gray!20}\multirow{-2}{*}{Method}&Frame Rate &Memory &I-AUROC \\ 
                \midrule[0.5pt]
            BTF &3.197 	&3810.6  &86.5  \\
            AST &4.966 	&4639.4  &93.7  \\
            M3DM &0.514 &65261.2  &\fcolorbox{firstcolor}{white}{94.5} \\
             IB-IUMAD &\fcolorbox{firstcolor}{white}{\textbf{21.427}}	&\fcolorbox{firstcolor}{white}{\textbf{1483.7}}	&\textbf{91.0}\\
            \bottomrule[1.0pt]
    \end{tabular}}}}
\end{table}

\section{Related Work}
\textbf{N-objects-N-models} is a typical anomaly detection para- digm, which refers to training a separate model for each object to identify anomalies \cite{defard2021padim,jiang2022softpatch,wang2025distribution,yu2025background}. Specifically, early efforts primarily focused on flaw inspection in RGB images, utilizing feature embedding \cite{li2021cutpaste,rudolph2022fully, deng2022anomaly, bergmann2020uninformed,li2024promptad,liang2025look,du20253d} and reconstruction-based methods \cite{schluter2022natural,zavrtanik2021draem,zavrtanik2022dsr,zavrtanik2024cheating,fang2025boosting,zhang2025costfilter}.  Recently, with the emergence of multimodal anomaly detection (MAD) \cite{horwitz2023back,asymmetric,li2024towards,zhao2025unimmad,tu2024self}, which leverages the complementary information of RGB and depth images to enhance detection performance, has attracted considerable attention \cite{chu2023shape}. For example, methods like M3DM \cite{wang2023multimodal}, CFM \cite{costanzino2024multimodal}, and FIND \cite{li2025find} have substantially advanced defect detection through different fusion strategies. However, despite its near-saturated performance in single-category tasks, the N-objects-N-models paradigm faces substantial challenges in scaling to multi-category anomaly detection.

\textbf{N-objects-one-model} paradigm is an emerging and promising approach that aims to train a single model capable of detecting anomalies across all objects \cite{dai2024seas,lu2023hierarchical,zhao2023omnial,lu2023removing,zhang2024learning,yao2024hierarchical}. A pioneering approach, UniAD \cite{you2022unified}, employs a Transformer-based architecture for feature reconstruction, sparking a wave of subsequent research \cite{yao2024prior,cheng2025boosting,chen2025center,wei2025uninet,zhang2023exploring}. Such as, DiAD \cite{he2024diffusion} introduces diffusion models into RGB-based multi-class anomaly detection via a semantics-guided network, while He \textit{et al.} \cite{DBLP:conf/nips/HeBZHCGWLT024} achieve efficient and lightweight flaw detection by integrating Mamba. However, existing studies have largely overlooked the development of unified MAD models. 

Moreover, in industrial scenarios, the frequent emergence of new objects requires MAD models within the N-objects-one-model framework to support incremental learning (IL) for effective adaptation. Although IL has shown promise in RGB-based anomaly detection (e.g., IUF \cite{tang2024incremental}, CDAD \cite{li2025one}), existing approaches \cite{lee2025continual,DBLP:conf/ijcai/HuGDLLHC25,lu2025c3d} often neglect the influence of spurious and redundant features, which exacerbate catastrophic forgetting during incremental updates. This issue becomes more serious in the MAD due to the inherent complexity of cross-modal fusion. To address this issue, we propose a novel incremental unified multimodal denoising framework for MAD.

\begin{figure}[t!]
 \centering
      \includegraphics[scale=0.31]{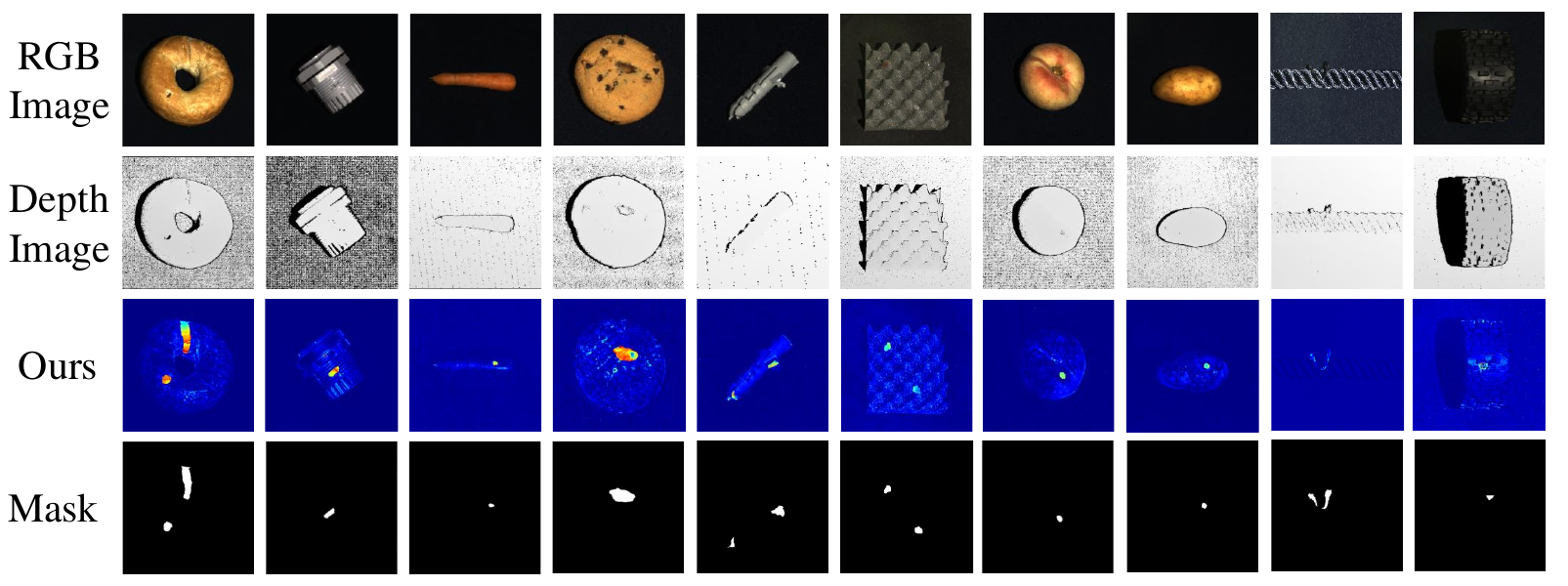}
  \caption{Visualizations results on MVTec 3D-AD.}
 \label{fig:54444}
\end{figure}
\section{Conclusion}
This paper investigates the impact of spurious and redundant features on catastrophic forgetting in IUMAD tasks. To address this challenge, we propose IB-IUMAD, an incremental unified multimodal denoising framework that integrates the complementary strengths of the Mamba decoder and IBFM module. This design effectively mitigates inter-object spurious feature interference and filters out redundant information in multimodal fusion features, thereby substantially alleviating catastrophic forgetting and achieving consistent performance gains. 
Extensive experiments on multiple datasets validate the effectiveness and superiority of our proposed method. 
We hope this work will inspire further research on MAD from an incrementally unified perspective, opening up new possibilities.
\section*{Acknowledgments}
This work is supported by the National Natural Science Foundation of China under Grant 62472079.
{
    \small
    \bibliographystyle{ieeenat_fullname}
    \bibliography{main}
}


\end{document}